\documentclass[10pt,twocolumn,letterpaper]{article}

\usepackage{iccv}
\usepackage{times}
\usepackage{epsfig}
\usepackage{graphicx}
\usepackage{amsmath}
\usepackage{amssymb}
\usepackage{tikz}
\usepackage{pgfplots}
\usepackage{stmaryrd}
\usepackage{booktabs}
\usepackage{bbm}
\usepackage{pdfpages}

 \makeatletter
 \DeclareRobustCommand\onedot{\futurelet\@let@token\@onedot}
 \def\@onedot{\ifx\@let@token.\else.\null\fi\xspace}
  
 \def\ie{i.e\onedot}

 \makeatother

\DeclareRobustCommand{\Figsref}[1]{Figures~\ref{#1}}

\DeclareRobustCommand{\Secref}[1]{Section~\ref{#1}}

\DeclareRobustCommand{\Eqnsref}[1]{Equations~(\ref{#1})}

\newcommand{\myparagraph}[1]{\vspace{-0.25cm} \paragraph{#1}}

\graphicspath{{./fig/}{./fig/plots/}}

\newcommand{\AproachName}{Neural-Image-QA\xspace}
\newcommand{\daquarNew}{DAQUAR-Consensus\xspace}

\usepackage[pagebackref=true,breaklinks=true,letterpaper=true,colorlinks,bookmarks=false]{hyperref}

\newcommand{\bs}[1]{\ensuremath{\boldsymbol{#1}}}

\DeclareMathOperator* {\argmax}{arg\, max}

\iccvfinalcopy %

\def\iccvPaperID{1679} %

\ificcvfinal\fi
\begin{document}

\title{Ask Your Neurons: A Neural-based Approach to\\Answering Questions about Images}

\newcommand{\authSpace}{&}

\author{Mateusz Malinowski$^{1}$ \and Marcus Rohrbach$^{2}$ \and  Mario Fritz$^{1}$ \and
$^{1}$Max Planck Institute for Informatics, Saarbr{\"u}cken, Germany \\
$^{2}$UC Berkeley EECS and ICSI, Berkeley, CA, United States
}

\maketitle
\begin{abstract}
We address a question answering task on real-world images that is set up as a Visual Turing Test. By combining latest advances in image representation and natural language processing, we propose \AproachName, an end-to-end formulation to this problem for which all parts are trained jointly. In contrast to previous efforts, we are facing a multi-modal problem where the language output (answer) is conditioned on visual and natural language input (image and question). Our approach \AproachName doubles the performance of the previous best approach on this problem. We provide additional insights into the problem by analyzing how much information is contained only in the language part for which we provide a new human baseline. 
To study human consensus, which is related to the ambiguities inherent in this challenging task, we propose two novel metrics and collect additional answers which extends the original DAQUAR dataset to \daquarNew. 
\end{abstract}

\section{Introduction}\label{sec:intro}

With the advances of natural language processing and image understanding, more complex and demanding tasks have become within reach. Our aim is to take advantage of the most recent developments to push the state-of-the-art for answering natural language questions on real-world images. This task unites inference of question intends and visual scene understanding with a word sequence prediction task.

Most recently, architectures based on the idea of layered, end-to-end trainable artificial neural networks have improved the state of the art across a wide range of diverse tasks. Most prominently Convolutional Neural Networks have raised the bar on image classification tasks \cite{krizhevsky2012imagenet} and Long Short Term Memory Networks are dominating performance on a range of sequence prediction tasks such as machine translation \cite{sutskever14nips}.

\begin{figure}[t]
\begin{center}
  \includegraphics[width=\columnwidth]{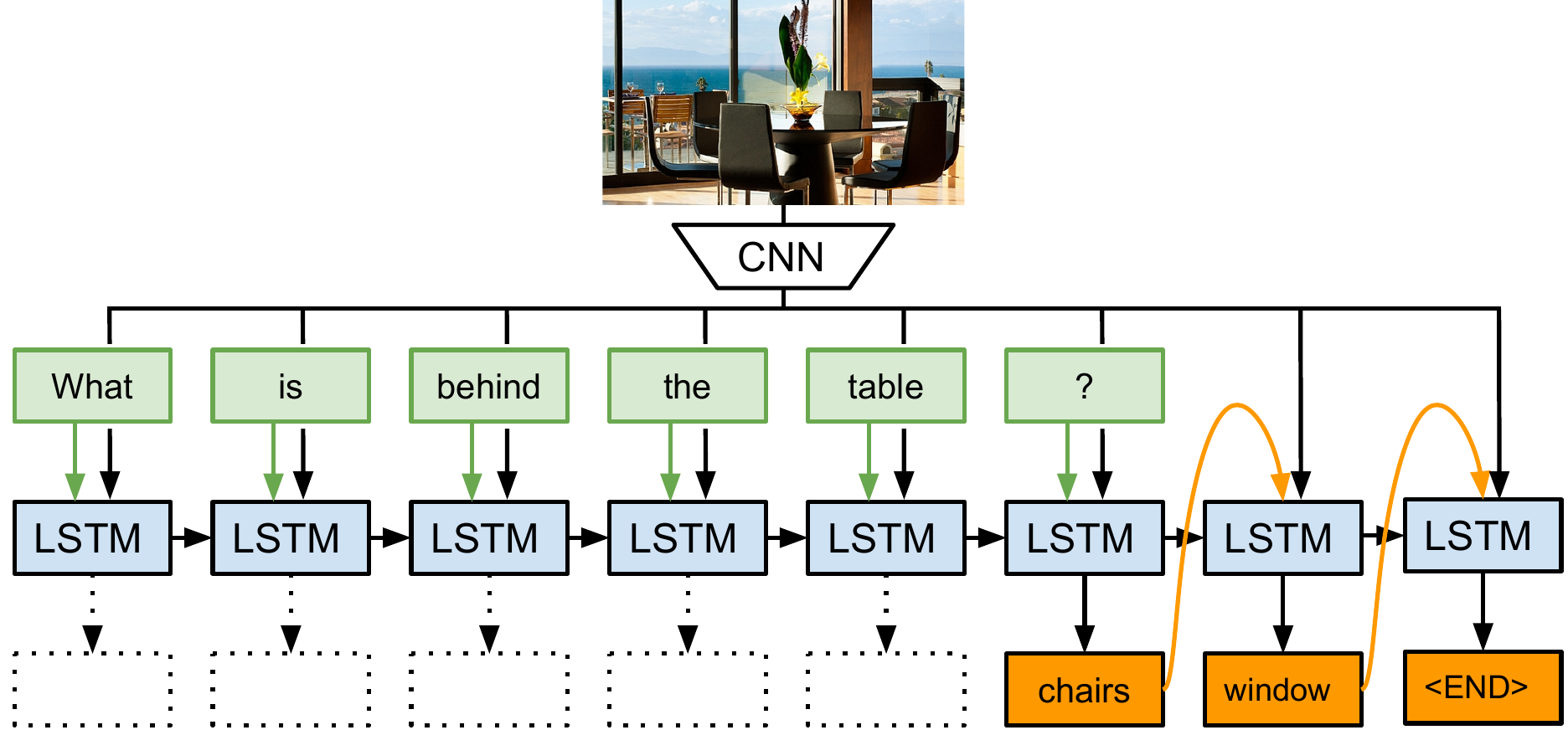}
  \caption[labelInTOC]{Our approach \AproachName to question answering with a Recurrent Neural Network using Long Short Term Memory (LSTM). To answer a question about an image, we feed in both, the image (CNN features) and the question (green boxes) into the LSTM. After the (variable length) question is encoded, we generate the answers (multiple words, orange boxes). During the answer generation phase the previously predicted answers are fed into the LSTM until the $\langle$END$\rangle$ symbol is predicted.}
  \label{fig:teaser}
\end{center}
\end{figure}

Very recently these two trends of employing neural architectures have been combined fruitfully with methods that can generate image \cite{karpathy15cvpr} and video descriptions \cite{venugopalan15iccv}. Both are conditioning on the visual features that stem from deep learning architectures and employ recurrent neural network approaches to produce descriptions.

To further push the boundaries and explore the limits of deep learning architectures, we propose an architecture for answering questions about images. In contrast to prior work, this task needs conditioning on language as well visual input. Both modalities have to be interpreted and jointly represented as an answer depends on inferred meaning of the question and image content.

While there is a rich body of work on natural language understanding that has addressed textual question answering tasks based on semantic parsing, symbolic representation and deduction systems, which also has seen applications to question answering on images \cite{malinowski14nips}, there is initial evidence that deep architectures can indeed achieve a similar goal
\cite{weston2014memory}.
This motivates our work to seek end-to-end architectures that learn to answer questions in a single holistic and monolithic model.

We propose \AproachName, an approach to question answering with a recurrent neural network.
An overview is given in \autoref{fig:teaser}. The image is analyzed via a Convolutional Neural Network (CNN) and the question together with the visual representation is fed into a Long Short Term Memory (LSTM) network.
The system is trained to produce the correct answer to the question on the image.
CNN and LSTM are trained jointly and end-to-end starting from words and pixels. 
\\
\noindent
\textbf{Contributions}:
We proposes a novel approach based on recurrent neural networks for the challenging task of answering of questions about images. It combines a CNN with a LSTM into an end-to-end architecture that predict answers conditioning on a question and an image. Our approach
significantly outperforms prior work on this task -- doubling the performance.
We collect additional data to study human consensus on this task, propose two new metrics sensitive to these effects, and provide a new baseline, by asking humans to answer the questions without observing the image.
We demonstrate a variant of our system that also answers question 
without accessing any visual information, which beats the human baseline. 
\section{Related Work}\label{sec:related_work}
As our method touches upon different areas in machine learning, computer vision  and natural language processing, we have organized related work in the following way:
\myparagraph{Convolutional Neural Networks for visual recognition.}
We are building on the recent success of Convolutional Neural Networks (CNN) for visual recognition \cite{krizhevsky2012imagenet,lecun1998gradient,ILSVRCarxiv14}, that are directly learnt from the raw image data and pre-trained on large image corpora.
Due to the rapid progress in this area within the last two years, a rich set of models \cite{simonyan2014very,szegedy2014going} is at our disposal.
\myparagraph{Recurrent Neural Networks (RNN) for sequence modeling.}
Recurrent Neural Networks
allow Neural Networks to handle
sequences of flexible length. A particular variant called Long Short Term Memory (LSTM) \cite{hochreiter97nc} has shown recent success on natural language tasks such as machine translation \cite{cho2014learning,sutskever14nips}.

\myparagraph{Combining RNNs and CNNs for description of visual content.}
The task of describing visual content like still images as well as videos has been successfully addressed with a combination of the previous two ideas \cite{donahue15cvpr,karpathy15cvpr,venugopalan15naacl,vinyals2014show,zitnick2013learning}. This is achieved by using the RNN-type model that first gets to observe the visual content and is trained to afterwards predict a sequence of words that is a description of the visual content. Our work extends this idea to question answering, where we formulate a model trained to generate an answer based on visual as well as natural language input.

\myparagraph{Grounding of natural language and visual concepts.}
Dealing with natural language input does involve the association of words with meaning. This is often referred to as grounding problem - in particular if the ``meaning'' is associated with a sensory input. While such problems have been historically addressed by symbolic semantic parsing techniques \cite{krishnamurthy2013jointly,matuszek2012joint}, there is a recent trend of machine learning-based approaches \cite{karpathy15cvpr,karpathy2014deep,kong2014you}
to find the associations.
Our approach follows the idea that we do not enforce or evaluate any particular representation of ``meaning'' on the language or image modality. We treat this as latent and leave this to the joint training approach to establish an appropriate internal representation for the question answering task.

\myparagraph{Textual question answering.}
Answering on purely textual questions has been studied in the NLP community \cite{berant2014semantic,liang2013learning} and state of the art techniques typically employ semantic parsing to arrive at a logical form capturing the intended meaning and infer relevant answers. Only very recently, the success of the previously mentioned neural sequence models as RNNs has carried over to this task \cite{iyyer2014neural,weston2014memory}.
More specifically \cite{iyyer2014neural} uses dependency-tree Recursive NN instead of LSTM, and reduce the question-answering problem to a classification task. Moreover, according to \cite{iyyer2014neural} their method cannot be easily applied to vision. \cite{weston2014memory} propose different kind of network - memory networks - and it is unclear how to apply \cite{weston2014memory} to take advantage of the visual content. However, neither \cite{iyyer2014neural} nor \cite{weston2014memory} show an end-to-end, monolithic approaches that produce multiple words answers for question on images.

\myparagraph{Visual Turing Test.} %
Most recently several approaches have been proposed to approach Visual Turing Test \cite{malinowski14visualturing}, \ie answering question about visual content.
For instance \cite{geman2015visual} have proposed a binary (yes/no) version of Visual Turing Test on synthetic data.
  In \cite{malinowski14nips}, we present a question answering system based on a semantic parser on a more varied set of human question-answer pairs.
  In contrast, in this work, our method is based on a neural architecture, which is trained end-to-end and therefore liberates the approach from any ontological commitment that would otherwise be introduced by a semantic parser.

We like to note that shortly after this work, several neural-based models \cite{ren2015image,learning_to_answer_questions,gao2015you}
have also been suggested.
Also several new datasets for
Visual Turing Tests
have just been proposed \cite{antol2015vqa,visual_madlibs15} that are worth further investigations.

\section{Approach}\label{sec:method}

Answering questions on images is the problem of predicting an answer %
$\bs{a}$ given an image $\bs{x}$ and a question $\bs{q}$ according to a parametric probability measure:
\begin{equation}
\bs{\hat{a}}=\argmax_{\bs{a}\in \mathcal{A}}p(\bs{a}|\bs{x},\bs{q};\bs{\theta})
\end{equation}
where $\bs{\theta}$ represent a vector of all parameters to learn and $\mathcal{A}$ is a set of all answers. %
Later we describe how we represent $\bs{x}$, $\bs{a}$, $\bs{q}$, and $p(\cdot|\bs{x},\bs{q};\bs{\theta})$ in more details.
In our scenario questions can have multiple word answers and we consequently decompose the problem to predicting a set of answer words $\bs{a}_{\bs{q},\bs{x}} = \left\{\bs{a}_1, \bs{a}_2, ..., \bs{a}_{\mathcal{N}(q,x)}\right\}$, where $\bs{a}_t$ are words from a finite vocabulary $\mathcal{V'}$, and $\mathcal{N}(q,x)$ is the number of answer words for the given question and image.
In our approach, named \AproachName, we propose to tackle the problem as follows.  To predict multiple words we formulate the problem as predicting a sequence of words from the vocabulary $\mathcal{V}:=\mathcal{V'}\cup\left\{\$\right\}$ where the extra token $\$$ indicates the end of the answer sequence, and points out that the question has been fully answered. %
We thus formulate the prediction procedure recursively:
\begin{equation}
\label{eq:recursivePred}
\bs{\hat{a}}_t=\argmax_{\bs{a}\in \mathcal{V}}p(\bs{a}|\bs{x},\bs{q},\hat{A}_{t-1};\bs{\theta})
\end{equation}
where $\hat{A}_{t-1}=\left\{\bs{\hat{a}}_1,\ldots,\bs{\hat{a}_{t-1}}\right\}$ is the set of previous words, with $\hat{A}_{0}=\left\{\right\}$ at the beginning, when our approach has not given any answer so far. The approach is terminated when $\hat{a}_t=\$$. 
We evaluate the method solely based on the predicted answer words ignoring the extra token $\$$.
To ensure uniqueness of the predicted answer words, as we want to predict the \emph{set} of answer words, the prediction procedure can be
be trivially changed  
by maximizing over
$\mathcal{V}\setminus\hat{A}_{t-1}$. However, in practice, our algorithm learns to not predict any previously predicted words.
\begin{figure}[t]
\begin{center}
  \includegraphics[width=.7\columnwidth]{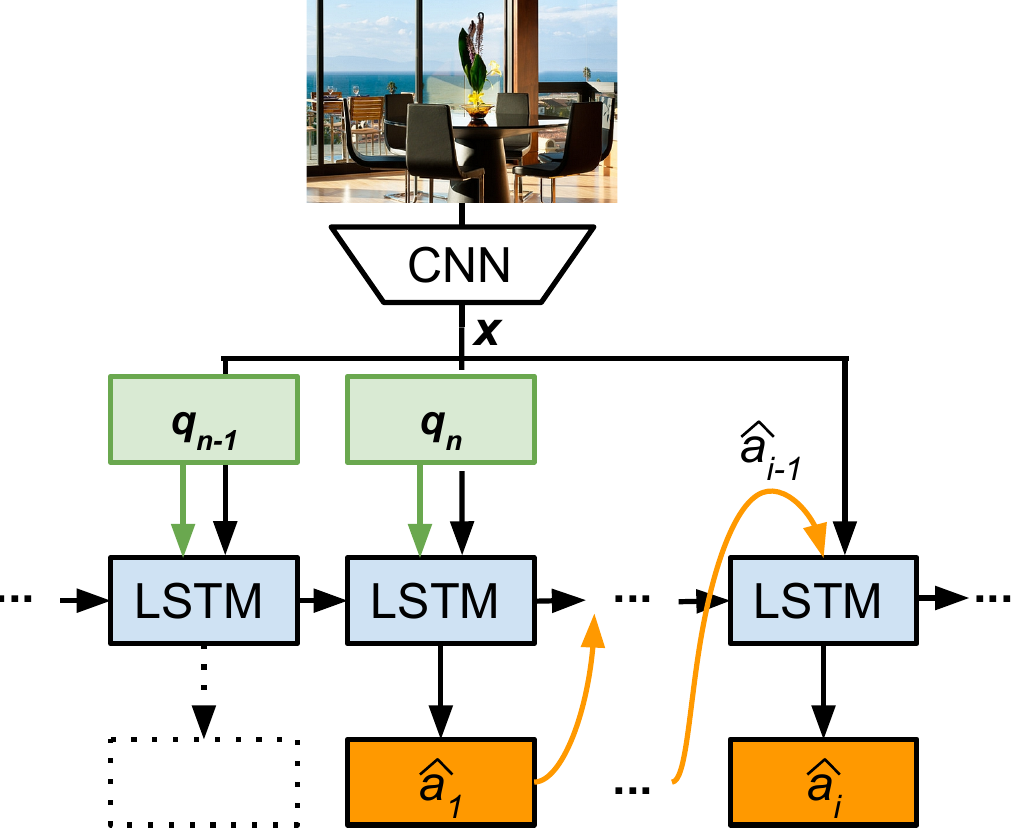}
  \caption{Our approach \AproachName,  see \Secref{sec:method} for details.}
  \label{fig:lstm-approach}
\end{center}
\end{figure}
\\
As shown in \autoref{fig:teaser} and \autoref{fig:lstm-approach}, we  feed \AproachName with a question as a sequence of words, \ie 
$\bs{q}=\left[\bs{q}_1,\ldots,\bs{q}_{n-1},\llbracket ? \rrbracket\right]$, where each $\bs{q}_t$ is the $t$-th word question  and $\llbracket ? \rrbracket:=\bs{q}_n$ encodes the question mark - the end of the question.
Since our problem is formulated as a variable-length input/output sequence,
we model the parametric distribution  $p(\cdot|\bs{x},\bs{q};\bs{\theta})$ of  \AproachName with a recurrent neural network and a  softmax prediction layer. More precisely, \AproachName is a deep network built of CNN \cite{lecun1998gradient} and Long-Short Term Memory (LSTM) \cite{hochreiter97nc}. LSTM has been recently shown to be effective in learning a variable-length sequence-to-sequence mapping \cite{donahue15cvpr,sutskever14nips}.

Both question and answer words are represented with one-hot vector encoding (a binary vector with exactly one non-zero entry at the position indicating the index of the word in the vocabulary) and embedded in a lower dimensional space, using a jointly learnt latent linear embedding.
In the training phase, we augment the question words sequence $\bs{q}$ with the corresponding ground truth answer words sequence $\bs{a}$, \ie $\bs{\hat{q}} := \left[\bs{q}, \bs{a}\right]$. During the test time, in the prediction phase, at time step $t$, we augment $\bs{q}$ with previously predicted answer words $\hat{\bs{a}}_{1..t} := \left[\bs{\hat{a}}_1,\ldots,\bs{\hat{a}}_{t-1}\right]$, \ie $\bs{\hat{q}}_{t} := \left[\bs{q},\bs{\hat{a}}_{1..t}\right]$.
This means the question $\bs{q}$ and the previous answers are encoded implicitly in the hidden states of the LSTM, while the latent hidden representation is learnt. We encode the image $\bs{x}$ using a CNN and provide it at every time step as input to the LSTM. 
We set the input 
$\bs{v}_t$
as a concatenation of 
$\left[\bs{x}, \bs{\hat{q}}_t\right]$.

\begin{figure}[t]
\begin{center}
  \includegraphics[width=.64\columnwidth]{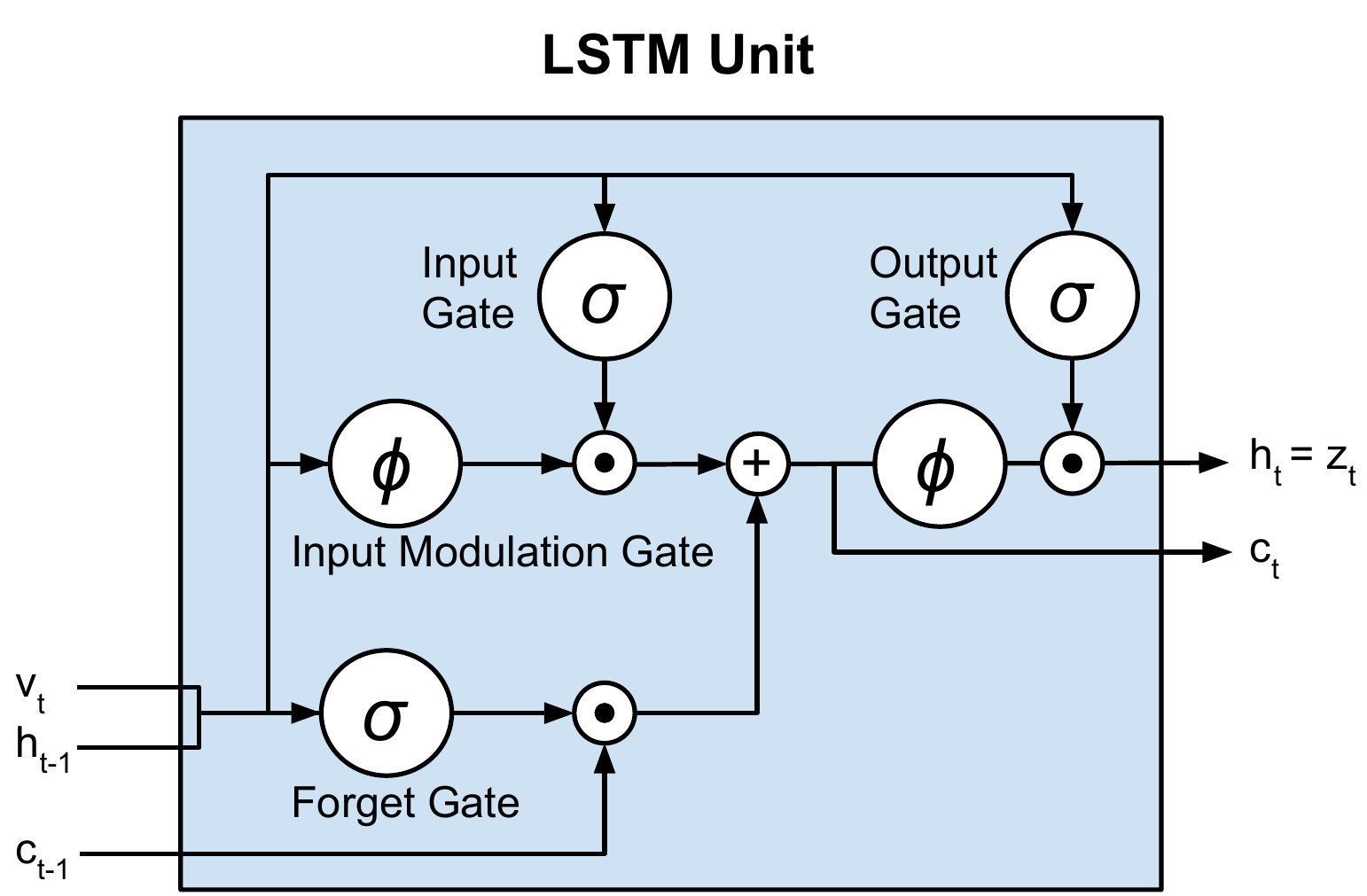}
  \caption[labelInTOC]{LSTM unit. See \Secref{sec:method}, \Eqnsref{eq:i}-(\ref{eq:h}) for details.}
  \label{fig:lstm}
\end{center}
\end{figure}
As visualized in detail in \autoref{fig:lstm}, the LSTM unit takes an input vector $\bs{v}_t$ at each time step $t$ and predicts an output 
word
 $\bs{z_t}$ which is equal to its latent hidden state
 $\bs{h}_t$. As discussed above $\bs{z_t}$ is a linear embedding of the corresponding answer word $\bs{a_t}$.
In contrast to a simple RNN unit the LSTM unit additionally maintains a memory cell $\bs{c}$. This allows to learn long-term dynamics more easily and significantly reduces the vanishing and exploding gradients problem~\cite{hochreiter97nc}.
More precisely, we use the LSTM unit as described in \cite{zaremba14arxiv} and the  \textit{Caffe} implementation from \cite{donahue15cvpr}.
 With the  \textit{sigmoid} nonlinearity $\sigma:\mathbb{R} \mapsto [0, 1]$, $\sigma(v) = \left(1 + e^{-v}\right)^{-1}$ and the \textit{hyperbolic tangent} nonlinearity $\phi:\mathbb{R} \mapsto [-1, 1]$, $\phi(v) = \frac{e^v - e^{-v}}{e^v + e^{-v}} = 2\sigma(2v) - 1$, the LSTM updates for time step $t$ given inputs $\bs{v}_t$, $\bs{h}_{t-1}$, and the memory cell $\bs{c}_{t-1}$ as follows:
\begin{align}
\bs{i}_t &= \sigma(W_{vi}\bs{v}_t + W_{hi}\bs{h}_{t-1} + \bs{b}_i)\label{eq:i}\\
\bs{f}_t &= \sigma(W_{vf}\bs{v}_t + W_{hf}\bs{h}_{t-1} + \bs{b}_f)\label{eq:f} \\
\bs{o}_t &= \sigma(W_{vo}\bs{v}_t + W_{ho}\bs{h}_{t-1} + \bs{b}_o) \label{eq:o}\\
\bs{g}_t &=   \phi(W_{vg}\bs{v}_t + W_{hg}\bs{h}_{t-1} + \bs{b}_g)\label{eq:g} \\
\bs{c}_t &= \bs{f}_t \odot \bs{c}_{t-1} + \bs{i}_t \odot \bs{g}_t \label{eq:c}\\
\bs{h}_t &= \bs{o}_t \odot \phi(\bs{c}_t)\label{eq:h}
\end{align}
where $\odot$ denotes element-wise multiplication.
All the weights $W$ and biases $b$ of the network are learnt jointly with the cross-entropy loss. Conceptually, as shown in \autoref{fig:lstm},  \autoref{eq:i} corresponds to the input gate, \autoref{eq:g} the input modulation gate, and \autoref{eq:f} the forget gate, which determines how much to keep from the previous memory $c_{t-1}$ state.
As \Figsref{fig:teaser} and \ref{fig:lstm-approach} suggest, all the output predictions that occur before the question mark are excluded from the loss computation, so that the model is penalized solely based on the predicted answer words.

\myparagraph{Implementation}
We use default hyper-parameters of LSTM \cite{donahue15cvpr} and CNN \cite{jia2014caffe}. All CNN models are first pre-trained on the ImageNet dataset \cite{ILSVRCarxiv14}, and next we randomly initialize and train the last layer together with the LSTM network on the task. We find this step crucial in obtaining good results.
We have explored the use of a 2 layered LSTM model, but have consistently obtained worse performance.
In a pilot study, we have found that \textit{GoogleNet} architecture \cite{jia2014caffe,szegedy2014going} consistently outperforms the \textit{AlexNet} architecture \cite{jia2014caffe,krizhevsky2012imagenet} as a CNN model for our task and model. 
\section{Experiments}\label{sec:results}
In this section we benchmark our method on a task of answering questions about images. We compare different variants of our proposed model to prior work in Section \ref{sec:experiments:eval}. In addition, in Section \ref{sec:experiments:evalNoImg}, we analyze how well questions can be answered without using the image in order to gain an understanding of biases in form of prior knowledge and common sense.
 We provide a new human baseline for this task.
In Section \ref{sec:experiments:humanConsensus} we discuss ambiguities in the question answering tasks and analyze them further by introducing metrics that are sensitive to these phenomena. In particular, the WUPS score \cite{malinowski14nips} is extended to a consensus metric that considers multiple human answers. Additional results are available in the supplementary material and on the project webpage \footnote{\url{https://www.d2.mpi-inf.mpg.de/visual-turing-challenge}}.

\myparagraph{Experimental protocol}
We evaluate our approach on the DAQUAR dataset \cite{malinowski14nips} which provides $12,468$ human question answer pairs on images of indoor scenes \cite{silbermanECCV12} and follow the same evaluation protocol by providing results on  accuracy and the WUPS score at $\left\{0.9,0.0\right\}$.
We run experiments for the full dataset as well as their proposed reduced set that restricts the output space to only $37$ object categories and uses $25$ test images. In addition, we also evaluate the methods on different subsets of DAQUAR where only $1$, $2$, $3$ or $4$ word answers are present.

\begin{table}
\begin{center}
\begin{tabular}{lrrr}
\toprule
 & Accu- & WUPS & WUPS\\
 &racy&@0.9&@0.0\\
 \cmidrule(lr){1-1}\cmidrule(lr){2-4}
  Malinowski et al. \cite{malinowski14nips}  & $7.86$&$11.86$&$38.79$\\
  \cmidrule(lr){1-1}\cmidrule(lr){2-4}

\AproachName (ours)\\
\ - multiple words& $17.49$&$23.28$&$57.76$ \\
\ - single word& $\boldsymbol{19.43}$&$\boldsymbol{25.28}$&$\boldsymbol{62.00}$\\
Human answers \cite{malinowski14nips}  & $50.20$&$50.82$&$67.27$\\

 \cmidrule(lr){1-1}\cmidrule(lr){2-4}

  Language only (ours)\\ 
  \ - multiple words& $17.06$&$22.30$&$56.53$  \\
\ - single word& $17.15$&$22.80$&$58.42$ \\
 Human answers, no images & $7.34$ & $13.17$ & $35.56$ \\

\bottomrule
\end{tabular}
\end{center}
\caption{
Results on DAQUAR, all classes, single reference, in \%.
}
\label{table:full_daquar}
\end{table}

\myparagraph{WUPS scores}
We base our experiments as well as the consensus metrics on WUPS scores \cite{malinowski14nips}. The metric is a generalization of the accuracy measure that accounts for word-level ambiguities in the answer words. For instance `carton' and `box' can be associated with a similar concept, and hence models should not be strongly penalized for this type of mistakes. Formally:

\begin{align*}
  \textrm{WUPS}(A,T) = \frac{1}{N} \sum_{i=1}^N\min\{ \prod_{a \in A^i} \max_{t\in T^i} \mu(a, t) ,\; \\ \prod_{t \in T^i} \max_{a \in A^i} \mu(a, t)
  \}
\end{align*}
To embrace the aforementioned ambiguities, \cite{malinowski14nips} suggest using a thresholded taxonomy-based Wu-Palmer similarity \cite{wu1994verbs} for $\mu$. The smaller the threshold the more forgiving metric. As in \cite{malinowski14nips}, we report WUPS at two extremes, $0.0$ and $0.9$.

\subsection{Evaluation of \AproachName}
\label{sec:experiments:eval}
We start with the evaluation of our \AproachName on the full DAQUAR dataset in order to study different variants and training conditions. Afterwards we evaluate on the reduced DAQUAR for additional points of comparison to prior work.

\myparagraph{Results on full DAQUAR}
\autoref{table:full_daquar} shows the results of our \AproachName method on the full set (``multiple words'') with $653$ images and $5673$ question-answer pairs available at test time. In addition, we evaluate a variant that is trained to predict only a single word (``single word'') as well as a variant that does not use visual features (``Language only'').
In comparison to the prior work \cite{malinowski14nips} (shown in the first row in \autoref{table:full_daquar}), we observe strong improvements of over $9\%$ points in accuracy and over $11\%$ in the WUPS scores [second row in \autoref{table:full_daquar} that corresponds to ``multiple words'']. Note that, we achieve this improvement despite the fact that the only published number available for the comparison
 on the full set uses ground truth object annotations \cite{malinowski14nips} -- which puts our method at a disadvantage.
Further improvements are observed when we train only on a single word answer, which doubles the accuracy obtained in prior work.
We attribute this to a joint training of the language and visual representations
and the dataset bias, where about $90\%$ of the answers contain only a single word.

We further analyze this effect in \autoref{fig:exactly_n_words}, where we show performance of our approach (``multiple words'') in dependence on the number of words in the answer (truncated at 4 words due to the diminishing performance). The performance of the ``single word'' variants on the one-word subset
are shown as horizontal lines. Although accuracy drops rapidly for longer answers, our model is capable of producing a significant number of correct two words answers. The ``single word'' variants have an edge on the single answers and benefit from the dataset bias towards these type of answers. Quantitative results of the ``single word'' model on the one-word answers subset of DAQUAR are shown in  \autoref{table:subset_single_word}.
While we have made substantial progress compared to prior work, there is still a $30\%$ points margin to human accuracy and $25$ in WUPS score [``Human answers'' in \autoref{table:full_daquar}].

\begin{table}
\begin{center}
\begin{tabular}{lrrr}
\toprule
 & Accu- & WUPS & WUPS\\
 &racy&@0.9&@0.0\\
 \cmidrule(lr){1-1}\cmidrule(lr){2-4}

\AproachName (ours)& $21.67$&$27.99$&$65.11
$\\
 \cmidrule(lr){1-1}\cmidrule(lr){2-4}
  Language only (ours)& $19.13$ & $25.16$ & $61.51$ \\
\bottomrule
\end{tabular}
\end{center}
\caption{Results of the single word model on the one-word answers subset of DAQUAR, all classes, single reference, in $\%$.
}
\label{table:subset_single_word}
\end{table}

\pgfplotsset{
compat=1.5,
ymax=30
}
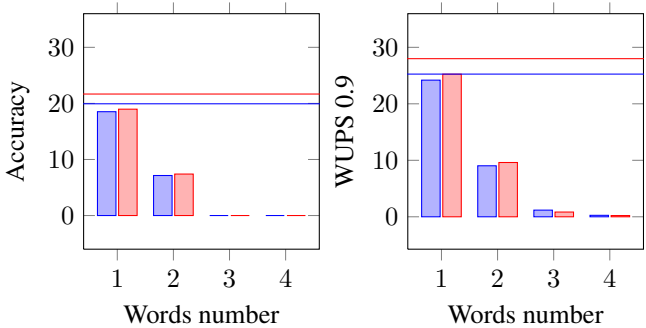
\begin{figure}
\hspace{-0.55cm}
\begin{tabular}{rl}
\begin{tikzpicture}
\begin{axis}[
  width=0.27\textwidth,
  height=0.27\textwidth,
	x tick label style={
		/pgf/number format/1000 sep=},
	ylabel=Accuracy,
  xlabel=Words number,
	enlargelimits=0.2,
  legend style={at={(0.5,-0.25)},
  anchor=north,legend columns=-1},
	ybar=1pt,
  bar width=7pt,
  xtick=data,
]
\addplot
	coordinates {(1,18.52) (2,7.14)
		 (3,0.0) (4,0.0)};

\addplot
	coordinates {(1,18.97) (2,7.40)
		(3,0.0) (4,0.0) };

\addplot[blue,sharp plot,update limits=false]
	coordinates {(0,19.93) (5,19.93)}
	node[below] at (axis cs:3,19.93) {};

\addplot[red,sharp plot,update limits=false]
	coordinates {(0,21.67) (5,21.67)}
	node[below] at (axis cs:3,21.67) {};

\end{axis}
\end{tikzpicture}
\begin{tikzpicture}
\begin{axis}[
  width=0.27\textwidth,
  height=0.27\textwidth,
	x tick label style={
		/pgf/number format/1000 sep=},
	ylabel=WUPS 0.9,
  xlabel=Words number,
	enlargelimits=0.2,
  legend style={at={(0.5,-0.25)},
    anchor=north,legend columns=-1},
	ybar=1pt,
  bar width=7pt,
  xtick=data,
]
\addplot
	coordinates {(1,24.19) (2,9.03)
		 (3,1.18) (4,0.26)};

\addplot
	coordinates {(1,25.25) (2,9.61)
		(3,0.84) (4,0.22) };

\addplot[blue,sharp plot,update limits=false]
	coordinates {(0,25.26) (5,25.26)}
	node[below] at (axis cs:3,25.26) {};

\addplot[red,sharp plot,update limits=false]
	coordinates {(0,27.99) (5,27.99)}
	node[below] at (axis cs:3,27.99) {};
\end{axis}
\end{tikzpicture}
\end{tabular}
\caption{Language only (blue bar) and \AproachName (red bar) ``multi word'' models evaluated on different subsets of DAQUAR. We consider $1$, $2$, $3$, $4$ word subsets. The blue and red horizontal lines represent ``single word'' variants evaluated on the answers with exactly $1$ word.}
\label{fig:exactly_n_words}
\end{figure}
\paragraph{Results on reduced DAQUAR}
In order to provide performance numbers that are comparable to the proposed Multi-World approach in \cite{malinowski14nips}, we also run our method on the reduced set with $37$ object classes and only $25$ images with $297$ question-answer pairs at test time.

\autoref{table:reduced_daquar} shows that \AproachName also improves on the reduced DAQUAR set, achieving $34.68\%$ Accuracy and $40.76\%$ WUPS at 0.9 substantially outperforming \cite{malinowski14nips} by  $21.95\%$ Accuracy and $22.6$ WUPS. Similarly to previous experiments, we achieve the best performance using the ``single word'' variant.

\subsection{Answering questions without looking at images}
\label{sec:experiments:evalNoImg}
In order to study how much information is already contained in questions, we train a version of our model that ignores the visual input.
The results are shown in \autoref{table:full_daquar} and \autoref{table:reduced_daquar} under ``Language only (ours)''.
The best ``Language only'' models with $17.15\%$ and $32.32\%$ compare very well in terms of accuracy to the best models that include vision. The latter achieve $19.43\%$ and $34.68\%$ on the full and reduced set respectively.

In order to further analyze this finding, we have collected a new human baseline ``Human answer, no image'', where we have asked participants to answer on the DAQUAR questions without looking at the images. It turns out that humans can guess the correct answer in
$7.86\%$
 of the cases by exploiting prior knowledge and common sense. Interestingly, our best ``language only'' model outperforms the human baseline by over $9\%$.
A substantial number of answers are plausible and resemble a form of common sense knowledge employed by humans to infer answers without having seen the image.

\subsection{Human Consensus}
\label{sec:experiments:humanConsensus}

\begin{table}
\begin{center}
\begin{tabular}{lrrr}
\toprule
 & Accu- & WUPS & WUPS\\
 &racy&@0.9&@0.0\\
 \cmidrule(lr){1-1}\cmidrule(lr){2-4}
  Malinowski et al. \cite{malinowski14nips}  & $12.73$&$18.10$&$51.47$\\
  \cmidrule(lr){1-1}\cmidrule(lr){2-4}
  \AproachName (ours)\\
\ - multiple words& $29.27$&$36.50$&$79.47$ \\
\ - single word& $\boldsymbol{34.68}$&$\boldsymbol{40.76}$&$79.54$\\
   \cmidrule(lr){1-1}\cmidrule(lr){2-4}
Language only (ours)\\ 
  \ - multiple words& $32.32$&$38.39$&$80.05$  \\
\ - single word& $31.65$&$38.35$&$\boldsymbol{80.08}$ \\
\bottomrule
\end{tabular}
\end{center}
\caption{
Results on reduced DAQUAR, single reference, with a reduced set of $37$ object classes and $25$ test images with $297$ question-answer pairs, in $\%$ 
}
\label{table:reduced_daquar}
\end{table}

We observe that in many cases there is an inter human agreement in the answers for a given image and question and this is also reflected by the human baseline performance on the question answering task of $50.20\%$ [``Human answers'' in \autoref{table:full_daquar}].
We study and analyze this effect further by extending our dataset to multiple human reference answers in \Secref{sec:extended_consensus_annotation}, and proposing a new measure -- inspired by the work in psychology
 \cite{cohen1960coefficient,fleiss1973equivalence,nakashole2013fine} -- that handles disagreement in \Secref{sec:consensus_measure}, as well as conducting additional experiments in \Secref{sec:consensus_results}.

\subsubsection{\daquarNew}\label{sec:extended_consensus_annotation}
In order to study the effects of consensus in the question answering task, we have asked multiple participants to answer the same question of the DAQUAR dataset given the respective image.
We follow the same scheme as in the original data collection effort, where the answer is a set of words or numbers. We do not impose any further restrictions on the answers.
This extends the original data \cite{malinowski14nips} to an average of $5$ test answers per image and question. We refer to this dataset as \daquarNew.

\subsubsection{Consensus Measures}\label{sec:consensus_measure}
While we have to acknowledge inherent ambiguities in our task, we seek a metric that prefers an answer that is commonly seen as preferred.
We make two proposals:

\paragraph{Average Consensus:}
We use our new annotation set that contains multiple answers per question in order to compute an expected score in the evaluation:
\begin{align}
\label{eq:consensus_metric}
\frac{1}{N K} \sum_{i=1}^N \sum_{k=1}^K \min\{ \prod_{a \in A^i} \max_{t\in T^i_k} \mu(a, t) ,\; \prod_{t \in T^i_k} \max_{a \in A^i} \mu(a, t)\}
\end{align}
where for the $i$-th question $A^i$ is the answer generated by the architecture and $T^i_k$ is the $k$-th possible human answer corresponding to the $k$-th interpretation of the question.
Both answers $A^i$ and $T^i_k$ are sets of the words, and $\mu$ is a membership measure, for instance WUP \cite{wu1994verbs}.
We call this metric
``Average Consensus Metric (ACM)'' since, in the limits, as $K$ approaches the total number of humans, we truly measure the inter human agreement of every question.

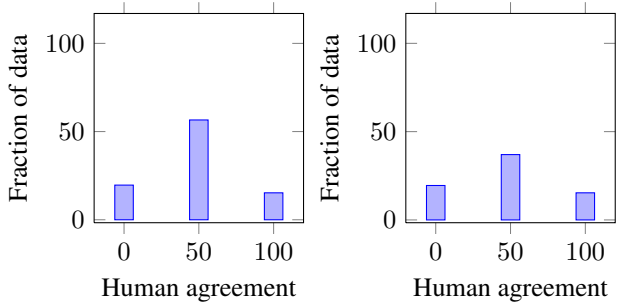
\begin{figure}
\begin{tabular}{rl}
\begin{tikzpicture}
\begin{axis}[
  width=0.25\textwidth,
  height=0.25\textwidth,
	x tick label style={
		/pgf/number format/1000 sep=},
  ylabel=Fraction of data,
  xlabel=Human agreement,
	enlargelimits=0.2,
  legend style={at={(0.5,-0.25)},
    anchor=north,legend columns=-1},
	ybar=1pt,
  bar width=7pt,
  xtick=data,
  ymax=100,
]
\addplot
  coordinates {(0,19.65) (50,56.58) (100, 15.27) };
\end{axis}
\end{tikzpicture}
\begin{tikzpicture}
\begin{axis}[
  width=0.25\textwidth,
  height=0.25\textwidth,
	x tick label style={
		/pgf/number format/1000 sep=},
  ylabel=Fraction of data,
  xlabel=Human agreement,
	enlargelimits=0.2,
  legend style={at={(0.5,-0.25)},
    anchor=north,legend columns=-1},
	ybar=1pt,
  bar width=7pt,
  xtick=data,
  ymax=100,
]
\addplot
  coordinates {(0,19.5) (50,37) (100, 15.35) };
\end{axis}
\end{tikzpicture}
\end{tabular}
\caption{Study of inter human agreement. At $x$-axis: no consensus ($0\%$), at least half consensus ($50\%$), full consensus ($100\%$). Results in $\%$. Left: consensus on the whole data, right: consensus on the test data.}
\label{fig:consensus_quality}
\end{figure}
\paragraph{Min Consensus:}
The Average Consensus Metric puts more weights on more ``mainstream'' answers due to the summation over possible answers given by humans. In order to measure if the result was at least with one human in agreement, we propose a ``Min Consensus Metric (MCM)'' by replacing the averaging in \autoref{eq:consensus_metric} with a max operator.
We call such metric Min Consensus and suggest using both metrics in the benchmarks. We will make the implementation of both metrics publicly available.
\begin{align}
\label{eq:interpretation_metric}
\frac{1}{N} \sum_{i=1}^N \max_{k=1}^K \left( \min\{ \prod_{a \in A^i} \max_{t\in T^i_k} \mu(a, t) ,\; \prod_{t \in T^i_k} \max_{a \in A^i} \mu(a, t)\} \right)
\end{align}
Intuitively, the max operator uses in evaluation a human answer that is the closest to the predicted one -- which represents a minimal form of consensus.

\begin{table}
\begin{center}
\begin{tabular}{lrrr}
\toprule
 & Accu- & WUPS & WUPS\\
 &racy&@0.9&@0.0\\
 \cmidrule(lr){1-1}\cmidrule(lr){2-4}
   {\bf Subset: No agreement }\\
     Language only (ours)\\ 
     \ - multiple words& $8.86$&$12.46$&$38.89$  \\
   \ - single word& $8.50$&$12.05$&$40.94$ \\
    \cmidrule(lr){1-1}\cmidrule(lr){2-4}
   \AproachName (ours)\\
   \ - multiple words& $\boldsymbol{10.31}$&$\boldsymbol{13.39}$&$40.05$ \\
   \ - single word&  $9.13$&$13.06$&$\boldsymbol{43.48} $
   \\
 \bottomrule
  {\bf Subset: $\ge 50\%$ agreement }\\
    Language only (ours)\\ 
    \ - multiple words& $21.17$&$27.43$&$66.68$  \\
  \ - single word& $20.73$&$27.38$&$67.69$ \\
   \cmidrule(lr){1-1}\cmidrule(lr){2-4}
  \AproachName (ours)\\
  \ - multiple words& $20.45$&$27.71$&$67.30$ \\
  \ - single word&  $\boldsymbol{24.10}$&$\boldsymbol{30.94}$&$\boldsymbol{71.95} $
  \\
  \bottomrule
   {\bf Subset: Full Agreement} \\
     Language only (ours)\\ 
     \ - multiple words& $27.86$&$35.26$&$78.83$  \\
   \ - single word& $25.26$&$32.89$&$79.08$ \\
    \cmidrule(lr){1-1}\cmidrule(lr){2-4}
   \AproachName (ours)\\
   \ - multiple words& $22.85$&$33.29$&$78.56$ \\
   \ - single word&  $\boldsymbol{29.62}$&$\boldsymbol{37.71}$&$\boldsymbol{82.31} $
   \\
   \bottomrule
\end{tabular}
\end{center}
\caption{
Results on DAQUAR, all classes, single reference in \% (the subsets are chosen based on \daquarNew).
}
\label{table:results_agreement_daquar}
\end{table}   

\subsubsection{Consensus results}\label{sec:consensus_results}
\begin{table}
\begin{center}
\begin{tabular}{lrrr}
\toprule
 & Accu- & WUPS & WUPS\\
 &racy&@0.9&@0.0\\
 \cmidrule(lr){1-1}\cmidrule(lr){2-4}
  {\bf Average Consensus Metric} \\
  Language only (ours)\\ 
  \ - multiple words& $11.60$&$18.24$&$52.68$  \\
\ - single word& $11.57$&$18.97$&$54.39$ \\
 \cmidrule(lr){1-1}\cmidrule(lr){2-4}
\AproachName (ours)\\
\ - multiple words& $11.31$&$18.62$&$53.21$ \\
\ - single word& $\boldsymbol{13.51}$&$\boldsymbol{21.36}$&$\boldsymbol{58.03} $
\\
\bottomrule
 {\bf Min Consensus Metric}\\
  Language only (ours)\\ 
  \ - multiple words& $22.14$&$29.43$&$66.88$  \\
\ - single word& $22.56$&$30.93$&$69.82$ \\
 \cmidrule(lr){1-1}\cmidrule(lr){2-4}
\AproachName (ours)\\
\ - multiple words& $22.74$&$30.54$&$68.17$ \\
\ - single word& $\boldsymbol{26.53}$&$\boldsymbol{34.87}$&$\boldsymbol{74.51} $\\
\bottomrule
\end{tabular}
\end{center}
\caption{
Results on \daquarNew, all classes, consensus in \%.
}
\label{table:aconsensus_daquar}
\end{table}

Using the multiple reference answers in \daquarNew we can show a more detailed analysis of inter human agreement. \autoref{fig:consensus_quality} shows the fraction of the data where the answers agree between all available questions (``100''), at least $50\%$ of the available questions and do not agree at all (no agreement - ``0''). We observe that for the majority of the data, there is a partial agreement, but even full disagreement is possible. We split the dataset into three parts according to the above criteria ``No agreement'', ``$\ge 50\%$ agreement'' and ``Full agreement'' and evaluate our models on these splits (\autoref{table:results_agreement_daquar} summarizes the results).
On subsets with stronger agreement, we achieve substantial gains of up to $10\%$ and $20\%$ points in accuracy over the full set (\autoref{table:full_daquar}) and the \textbf{Subset: No agreement} (\autoref{table:results_agreement_daquar}), respectively.
These splits can be seen as curated versions of DAQUAR, which allows studies with factored out ambiguities.

The aforementioned ``Average Consensus Metric'' generalizes the notion of the agreement, and encourages predictions of the most agreeable answers. On the other hand ``Min Consensus Metric'' has a desired effect of providing a more optimistic evaluation.  \autoref{table:aconsensus_daquar} shows the application of both measures to our data and models.

Moreover,  \autoref{table:consensus_human_baseline} shows that ``MCM'' applied to human answers at test time captures ambiguities in interpreting questions by improving the score of the human baseline from \cite{malinowski14nips}
(here, as opposed to \autoref{table:aconsensus_daquar}, we exclude the original human answers from the measure). It also cooperates well with WUPS at $0.9$, which takes word ambiguities into account, gaining an
 about $20\%$ higher score.%

\subsection{Qualitative results}

We show predicted answers of different variants of our architecture in \autoref{fig:vision_vs_language}, \ref{fig:multiple_answers}, and \ref{fig:mix_predictions}.
We have chosen the examples to highlight differences between \AproachName and the ``Language only''.
We use a ``multiple words'' approach only in \autoref{fig:multiple_answers}, otherwise the ``single word'' model is shown. Despite some failure cases, ``Language only'' makes ``reasonable guesses'' like predicting that the largest object could be table or an object that could be found on the bed is either a pillow or doll.

\begin{table}
\begin{center}
\begin{tabular}{lrrr}
\toprule
 & Accuracy & WUPS & WUPS\\
 &&@0.9&@0.0\\
 \cmidrule(lr){1-1}\cmidrule(lr){2-4}
  WUPS \cite{malinowski14nips}  & $50.20$&$50.82$&$67.27$\\
  \cmidrule(lr){1-1}\cmidrule(lr){2-4}
  ACM (ours) &$36.78$ & $45.68$ & $64.10$ \\
  MCM (ours) &$60.50$ & $69.65$ & $82.40$ \\
\bottomrule

\end{tabular}
\end{center}
\caption{
Min and Average Consensus on human answers from DAQUAR, as reference sentence we use all answers in \daquarNew which are not in DAQUAR, in $\%$ 
}
\label{table:consensus_human_baseline}
\end{table}
\subsection{Failure cases}
While our method answers correctly on a large part of the challenge (e.g. $\approx 35 $ WUPS at $0.9$ on ``what color'' and ``how many'' question subsets),  spatial relations ($\approx 21$ WUPS at $0.9$) which account for a substantial part of DAQUAR remain challenging.  Other errors involve questions with small objects, negations, and shapes (below $12$ WUPS at $0.9$). Too few training data points for the aforementioned cases may contribute to these mistakes.

\autoref{fig:mix_predictions} shows examples of failure cases that include (in order) strong occlusion, a possible answer not captured by our ground truth answers, and unusual instances (red toaster).
\begin{table*}[p!]
\begin{center}
\begin{tabular}{l@{\ }c@{\ }c@{\ }c}
\toprule
\multicolumn{2}{c}{\includegraphics[width=0.3\linewidth]{}} &
\includegraphics[width=0.3\linewidth]{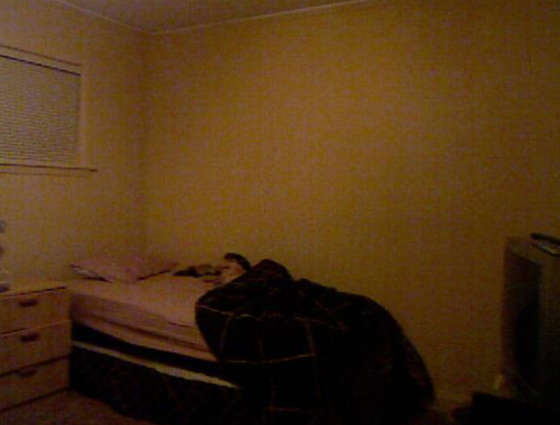} &
\includegraphics[width=0.3\linewidth]{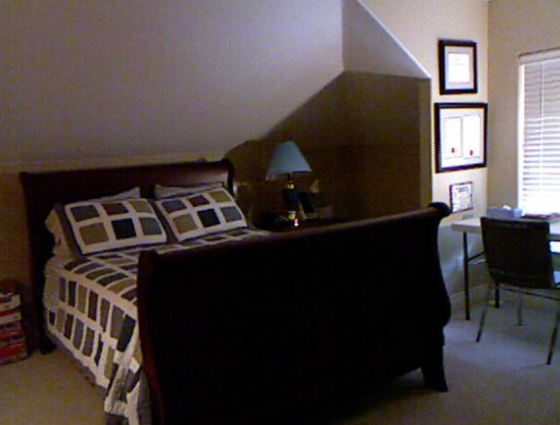} 
\\
\multicolumn{2}{c}{What is on the right side of the cabinet?} & \multicolumn{1}{c}{How many drawers are there?} & 
\multicolumn{1}{c}{What is the largest object?}
\\\midrule
\textit{\AproachName:}&\multicolumn{1}{c}{\textcolor{green}{bed}} & \multicolumn{1}{c}{\textcolor{green}{3}} & \multicolumn{1}{c}{\textcolor{green}{bed}}
\\\midrule
 \textit{Language only:} &\multicolumn{1}{c}{\textcolor{green}{bed}} & 
\multicolumn{1}{c}{\textcolor{red}{6}} & 
\multicolumn{1}{c}{\textcolor{red}{table}}\\
\bottomrule
\end{tabular}
\end{center}
\caption{Examples of questions and answers.  Correct predictions are colored in green, incorrect in red.}

\label{fig:vision_vs_language}
\end{table*}

\begin{table*}[tp]
\begin{center}
\begin{tabular}{l@{\ }c@{\ }c@{\ }c}
 \toprule
\multicolumn{2}{c}{\includegraphics[width=0.3\linewidth]{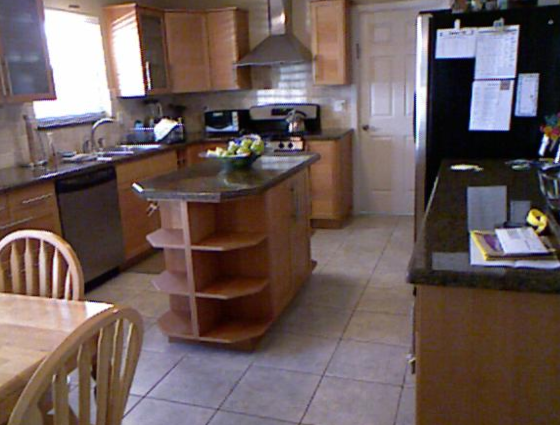}} &
\includegraphics[width=0.3\linewidth]{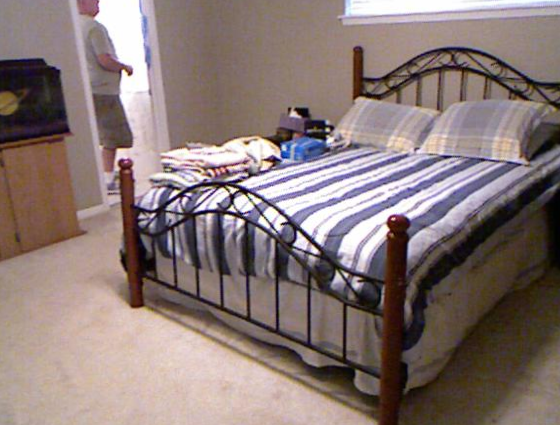} &
\includegraphics[width=0.3\linewidth]{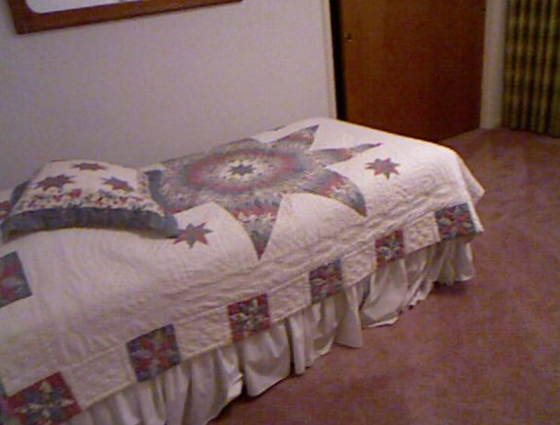} 
\\
\multicolumn{2}{c}{What is on the refrigerator?} & 
\multicolumn{1}{c}{What is the colour of the comforter?} & 
\multicolumn{1}{c}{What objects are found on the bed?}
\\\midrule
\textit{\AproachName:}&\multicolumn{1}{c}{\textcolor{green}{magnet, paper}} & 
\multicolumn{1}{c}{\textcolor{green}{blue, white}} & 
\multicolumn{1}{c}{\textcolor{green}{bed sheets, pillow}}
\\\midrule
\textit{Language only:}&\multicolumn{1}{c}{\textcolor{green}{magnet, paper}} & 
\multicolumn{1}{c}{\textcolor{red}{blue, green, red, yellow}} & 
\multicolumn{1}{c}{\textcolor{red}{doll, pillow}}\\
\bottomrule
\end{tabular}
\end{center}
\caption{Examples of questions and answers with multiple words. Correct predictions are colored in green, incorrect in red.
}

\label{fig:multiple_answers}
\end{table*}

\begin{table*}[tp]
\begin{center}
\begin{tabular}{l@{\ }c@{\ }c@{\ }c}
 \toprule
\multicolumn{2}{c}{\includegraphics[width=0.3\linewidth]{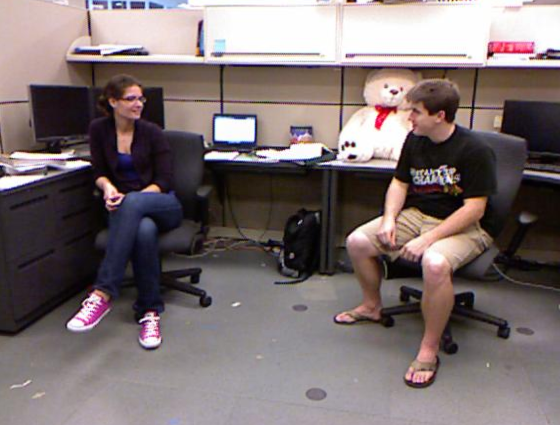}} &
\includegraphics[width=0.3\linewidth]{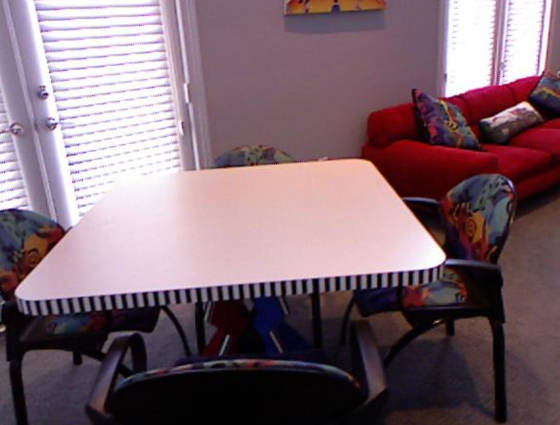} &
\includegraphics[width=0.3\linewidth]{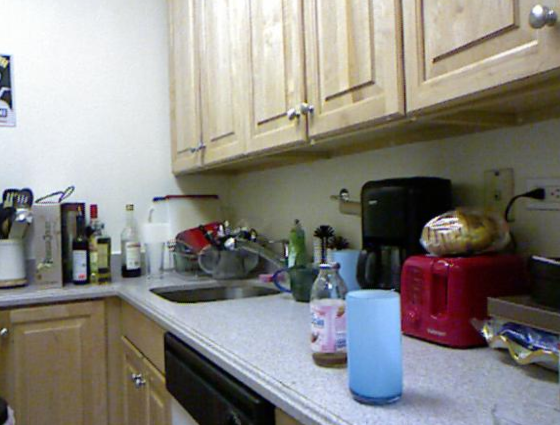} 
\\
\multicolumn{2}{c}{How many chairs are there?} & 
\multicolumn{1}{c}{What is the object fixed on the window?} & 
\multicolumn{1}{c}{Which item is red in colour?}
\\\midrule
\textit{\AproachName:}&\multicolumn{1}{c}{\textcolor{red}{1}} & 
\multicolumn{1}{c}{\textcolor{red}{curtain}} & 
\multicolumn{1}{c}{\textcolor{red}{remote control}}
\\\midrule
\textit{Language only:}&\multicolumn{1}{c}{\textcolor{red}{4}} & 
\multicolumn{1}{c}{\textcolor{red}{curtain}} & 
\multicolumn{1}{c}{\textcolor{red}{clock}}
\\\midrule
\textit{Ground truth answers:}&\multicolumn{1}{c}{\textcolor{black}{2}} & 
\multicolumn{1}{c}{\textcolor{black}{handle}} & 
\multicolumn{1}{c}{\textcolor{black}{toaster}}\\
\bottomrule
\end{tabular}
\end{center}
\caption{Examples of questions and answers - failure cases.}

\label{fig:mix_predictions}
\end{table*}

\section{Conclusions}\label{sec:conclusions}
We have presented a neural architecture for answering natural language questions about images that contrasts with prior efforts based on semantic parsing and outperforms prior work by doubling performance on this challenging task. A variant of our model that does not use the image to answer the question performs only slightly worse and even outperforms a new human baseline that we have collected under the same condition. We conclude that our model has learnt biases and patterns that can be seen as forms of common sense and prior knowledge that humans use to accomplish this task. We observe that indoor scene statistics, spatial reasoning, and small objects are not well captured by the global CNN representation, but the true limitations of this representation can only be explored on larger datasets. We extended our existing DAQUAR dataset to \daquarNew, which now provides multiple reference answers which allows to study inter-human agreement and consensus on the question answer task. We propose two new metrics: ``Average Consensus'', which takes into account human disagreement, and ``Min Consensus''
that captures disagreement in human question answering.

\paragraph{Acknowledgements.}
Marcus Rohrbach was supported by a fellowship within the FITweltweit-Program of the German Academic Exchange Service (DAAD).
\clearpage

\small
\bibliographystyle{ieee}
\bibliography{biblioShort,egbib,rohrbach,mario}



\section*{Supplementary material}

# Ask Your Neurons: A Neural-based Approach to Answering Questions about Images - A Supplementary Material

Mateusz Malinowski[1]   Marcus Rohrbach[2]   Mario Fritz[1]

[1]Max Planck Institute for Informatics, Saarbrücken, Germany

[2]UC Berkeley EECS and ICSI, Berkeley, CA, United States

|  | Accu-racy | WUPS @0.9 | WUPS @0.0 |
|---|---|---|---|
| Malinowski et al. [1] | 7.86 | 11.86 | 38.79 |
| Neural-Image-QA (ours) |  |  |  |
| - multiple words | 17.49 | 23.28 | 57.76 |
| - single word | **19.43** | **25.28** | **62.00** |
| Human answers [1] | 50.20 | 50.82 | 67.27 |
| Language only (ours) |  |  |  |
| - multiple words | 17.06 | 22.30 | 56.53 |
| - single word | 17.15 | 22.80 | 58.42 |
| Human answers, no images | 7.34 | 13.17 | 35.56 |

Table 1. Results on DAQUAR, all classes, single reference, in %. Replication of Table 1 from the main paper for convenience.

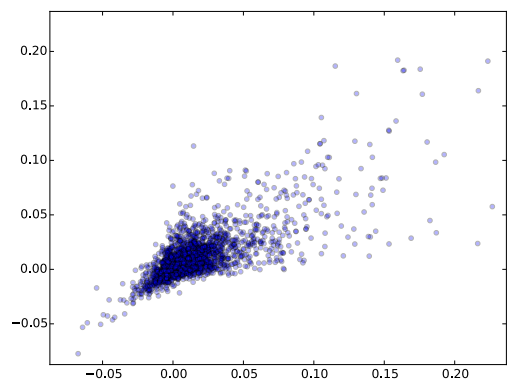

Figure 1. Figure showing correlation between question and answer words of the "Language only" model (at x-axis), and a similar correlation of the "Human-baseline" [1] (at y-axis).

In this supplemental material, we additionally provide qualitative examples of different variants of our architecture and show the correlations of predicted answer words and question words with human answer and question words.

The examples are chosen to highlight challenges as well as differences between "Neural-Image-QA" and "Language only" architectures. Table 9 also shows a few failure cases. In all cases but "multiple words answer", we use the best "single word" variants. Although "Language only" ignores the image, it is still able to make "reasonable guesses" by exploiting biases captured by the dataset that can be viewed as a type of common sense knowledge. For instance, "tea kettle" often sits on the oven, cabinets are usually "brown", "chair" is typically placed in front of a table, and we commonly keep a "photo" on a cabinet (Table 2, 4, 5, 8). This effect is analysed in Figure 1. Each data point in the plot represents the correlation between a question and a predicted answer words for our "Language only" model (x-axis) versus the correlation in the human answers (y-axis).

Despite the reasonable guesses of the "Language only" architecture, the "Neural-Image-QA" predicts in average better answers (shown in Table 1 that we have replicated from the main paper for the convenience of the reader) by exploiting the visual content of images. For instance in Table 6 the "Language only" model incorrectly answers "6" on the question "How many burner knobs are there ?" because it has seen only this answer during the training with exactly the same question but on different image.

Both models, "Language only" and "Neural-Image-QA", have difficulties to answer correctly on long questions or such questions that expect a larger number of answer words (Table 9). On the other hand both models are doing well on predicting a type of the question (e.g. "what color ..." result in a color name in the answer, or "how many ..." questions result in a number), there are a few rare cases with an incorrect type of the predicted answer (the last example in Table 9).

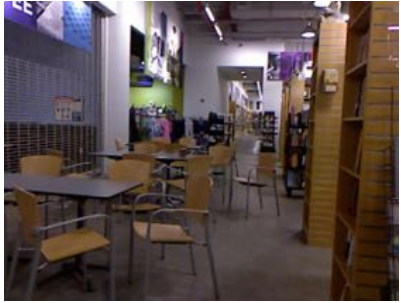 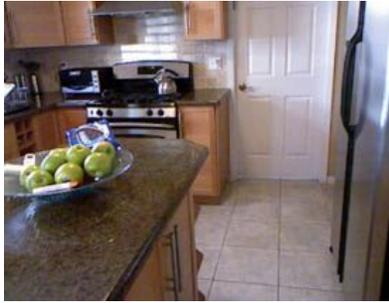 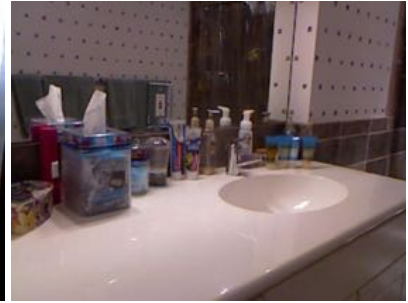

| | What are the objects close to the wall? | What is on the stove? | What is left of sink? |
|---|---|---|---|
| *Neural-Image-QA:* | wall decoration | tea kettle | tissue roll |
| *Language only:* | books | tea kettle | towel |
| *Ground truth answers:* | wall decoration | tea kettle | tissue roll |

Table 2. Examples of compound answer words.

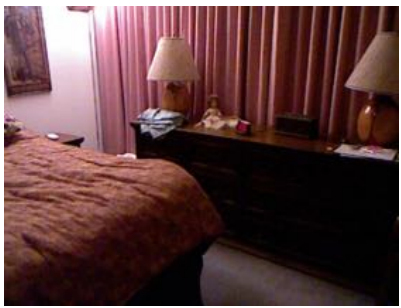 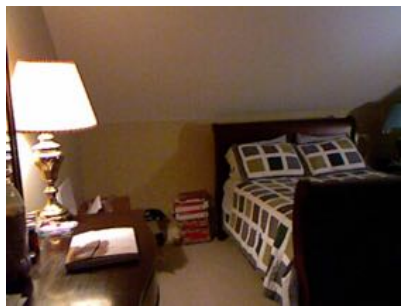 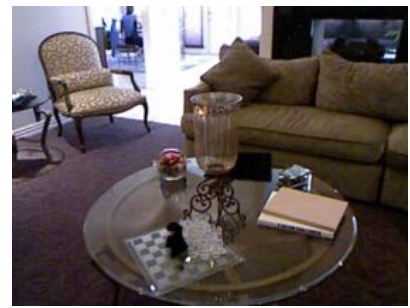

| | How many lamps are there? | How many pillows are there on the bed? | How many pillows are there on the sofa? |
|---|---|---|---|
| *Neural-Image-QA:* | 2 | 2 | 3 |
| *Language only:* | 2 | 3 | 3 |
| *Ground truth answers:* | 2 | 2 | 3 |

Table 3. Counting questions.

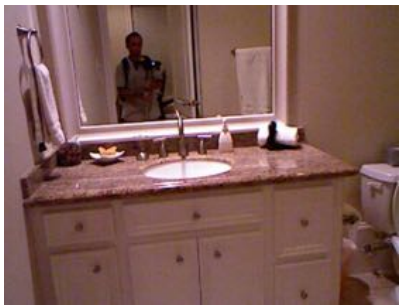 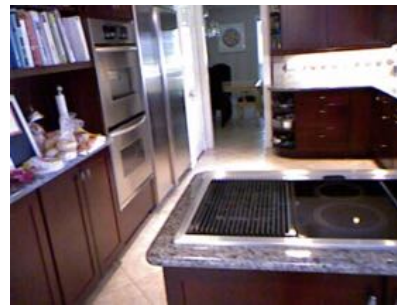 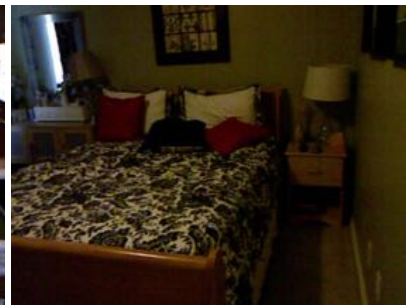

| | What color is the towel? | What color are the cabinets? | What is the colour of the pillows? |
|---|---|---|---|
| *Neural-Image-QA:* | brown | brown | black, white |
| *Language only:* | white | brown | blue, green, red |
| *Ground truth answers:* | white | brown | black, red, white |

Table 4. Questions about color.

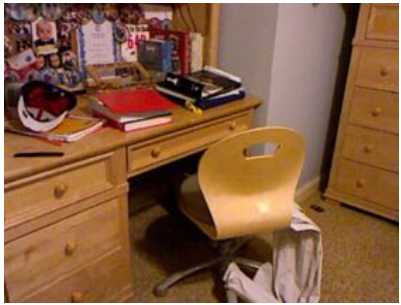 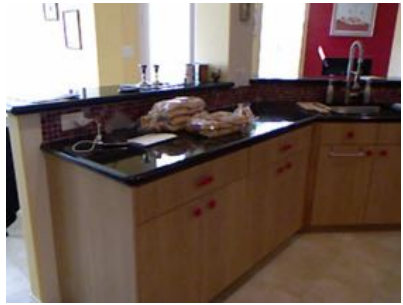 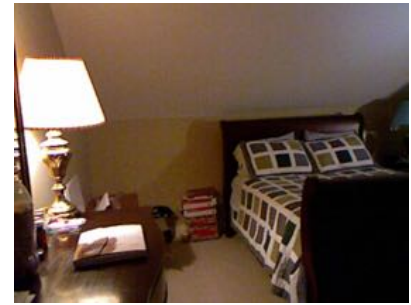

| | What is hanged on the chair? | What is the object close to the sink? | What is the object on the table in the corner? |
|---|---|---|---|
| *Neural-Image-QA:* | clothes | faucet | lamp |
| *Language only:* | jacket | faucet | plant |
| *Ground truth answers:* | clothes | faucet | lamp |

Table 5. Correct answers by our "Neural-Image-QA" architecture.

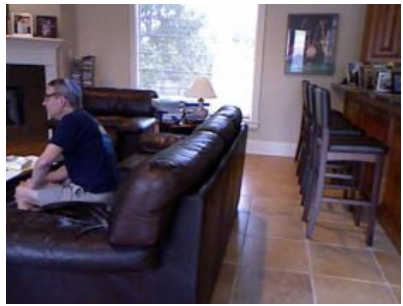 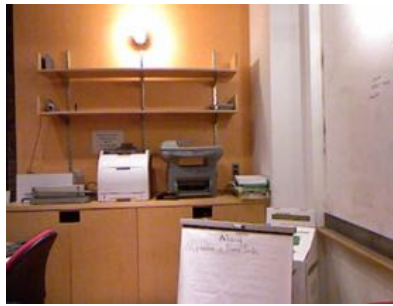 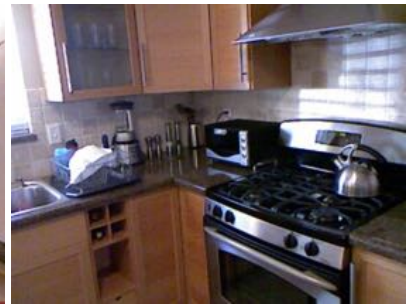

| | What are the things on the cabinet? | What is in front of the shelf? | How many burner knobs are there? |
|---|---|---|---|
| *Neural-Image-QA:* | photo | chair | 4 |
| *Language only:* | photo | basket | 6 |
| *Ground truth answers:* | photo | chair | 4 |

Table 6. Correct answers by our "Neural-Image-QA" architecture.

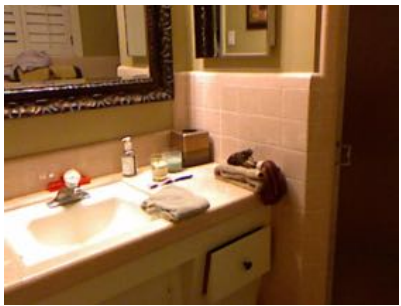 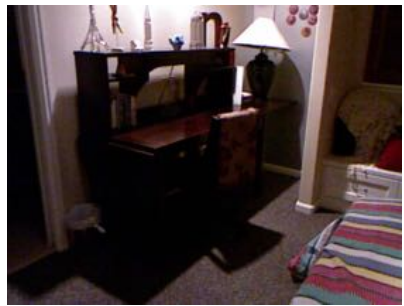 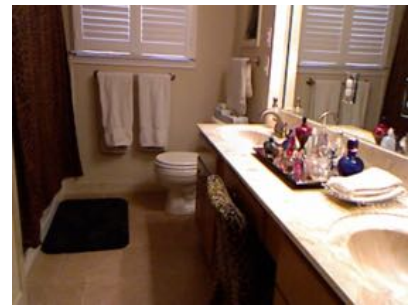

| | What is the object close to the counter? | What is the colour of the table and chair? | How many towels are hanged? |
|---|---|---|---|
| *Neural-Image-QA:* | sink | brown | 3 |
| *Language only:* | stove | brown | 4 |
| *Ground truth answers:* | sink | brown | 3 |

Table 7. Correct answers by our "Neural-Image-QA" architecture.

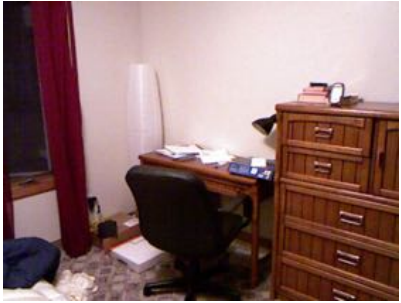 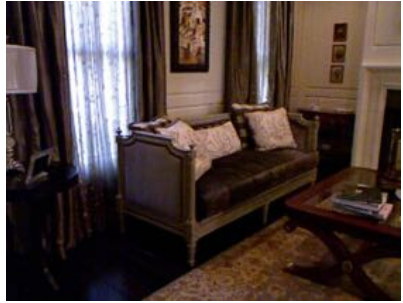 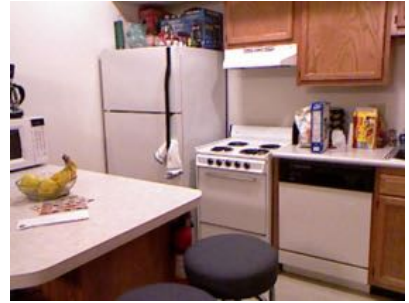

| | What is on the right most side on the table? | What are the things on the coffee table? | What is in front of the table? |
|---|---|---|---|
| *Neural-Image-QA:* | lamp | books | chair |
| *Language only:* | machine | jacket | chair |
| *Ground truth answers:* | lamp | books | chair |

Table 8. Correct answers by our "Neural-Image-QA" architecture.

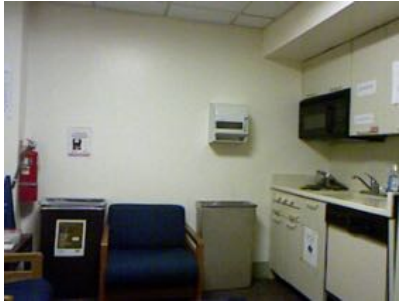 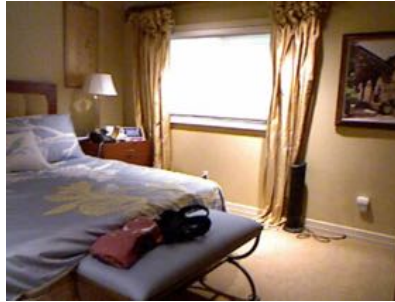 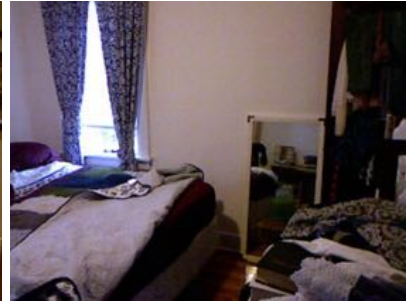

| | What is on the left side of the white oven on the floor and on right side of the blue armchair? | What are the things on the cabinet? | What color is the frame of the mirror close to the wardrobe? |
|---|---|---|---|
| *Neural-Image-QA:* | oven | chair, lamp, photo | pink |
| *Language only:* | exercise equipment | candelabra | curtain |
| *Ground truth answers:* | garbage bin | lamp, photo, telephone | white |

Table 9. Failure cases.



\end{document}